\documentclass[%
 aip,
 amsmath,amssymb,
 reprint,%
]{revtex4-1}

\usepackage{graphicx}% Include figure files
\usepackage{dcolumn}% Align table columns on decimal point
\usepackage{bm}% bold math
\usepackage{booktabs} % For better looking tables
\usepackage{tabularx} % For flexible-width columns For better control over captions
\usepackage[utf8]{inputenc}
\usepackage[T1]{fontenc}
\usepackage{mathptmx}
\usepackage{etoolbox}
\usepackage[inkscapelatex=false]{svg}
\usepackage{rotating} % Required for sidewaysfigure
\usepackage{adjustbox}

\makeatletter
\def\@email#1#2{%
 \endgroup
 \patchcmd{\titleblock@produce}
  {\frontmatter@RRAPformat}
  {\frontmatter@RRAPformat{\produce@RRAP{*#1\href{mailto:#2}{#2}}}\frontmatter@RRAPformat}
  {}{}
}%
\makeatother
\begin{document}

\preprint{AIP/123-QED}

\title{Adaptive control of recurrent neural networks using conceptors}

\author{Guillaume Pourcel}
\email{g.a.pourcel@rug.nl}

\altaffiliation{These authors contributed equally to this work.}
% \altaffiliation[Electronic mail: ]{g.a.pourcel@rug.nl}
\affiliation{ 
Bernoulli Institute for Mathematics, Computer Science and Artificial Intelligence \& Cognitive Systems
and Materials Center (CogniGron), University of Groningen, The Netherlands
}

\author{Mirko Goldmann}
\email{mirko@ifisc.uib-csic.es}
\altaffiliation{These authors contributed equally to this work.}
% \altaffiliation[Electronic mail: ]{mirko@ifisc.uib-csic.es}
\affiliation{ 
Instituto de F\'{i}sica Interdisciplinar y Sistemas Complejos (IFISC (UIB-CSIC)), Campus Universitat de les Illes Balears E-07122, Palma de Mallorca, Spain
}

\author{Ingo Fischer}
\affiliation{
Instituto de F\'{i}sica Interdisciplinar y Sistemas Complejos (IFISC (UIB-CSIC)), Campus Universitat de les Illes Balears E-07122, Palma de Mallorca, Spain
}

\author{Miguel C. Soriano}
\affiliation{
Instituto de F\'{i}sica Interdisciplinar y Sistemas Complejos (IFISC (UIB-CSIC)), Campus Universitat de les Illes Balears E-07122, Palma de Mallorca, Spain
}

\date{\today}

\begin{abstract}
Recurrent Neural Networks excel at predicting and generating complex high-dimensional temporal patterns. Due to their inherent nonlinear dynamics and memory, they can learn unbounded temporal dependencies from data. In a Machine Learning setting, the network’s parameters are adapted during a training phase to match the requirements of a given task/problem increasing its computational capabilities. After the training, the network parameters are kept fixed to exploit the learned computations. The static parameters thereby render the network unadaptive to changing conditions, such as external or internal perturbation. In this manuscript, we demonstrate how keeping parts of the network adaptive even after the training enhances its functionality and robustness. Here, we utilize the conceptor framework and conceptualize an adaptive control loop analyzing the network's behavior continuously and adjusting its time-varying internal representation to follow a desired target. We demonstrate how the added adaptivity of the network supports the computational functionality in three distinct tasks: interpolation of temporal patterns, stabilization against partial network degradation, and robustness against input distortion. Our results highlight the potential of adaptive networks in machine learning beyond training, enabling them to not only learn complex patterns but also dynamically adjust to changing environments, ultimately broadening their applicability.
\end{abstract}

\maketitle

\begin{quotation}
    Recurrent neural networks (RNNs) can capture complex temporal dependencies, making them an attractive choice for storing, retrieving, and predicting time series data. This ability renders them attractive models for solving machine learning tasks that require exploiting nonlinear and memory-based processing. During a training phase the parameters of a RNN are adapted based on a learning algorithm such as reservoir computing or backpropagation through time. Since the parameters of the network change in that phase, RNNs can be seen as adaptive networks during training. However, after the training, in the so-called inference phase, the parameters of the network are usually kept fixed. While this allows to exploit the learned dynamical behavior, it often renders the network unable to adapt to new and changing conditions. In our work, we propose a novel mechanism that keeps certain parameters of the network adaptive even in the inference. . In essence, conceptors control network dynamics by projecting them into a small subspace of network's state space that can be interpreted as an ellipsoid. Based on the conceptor framework, we design a control mechanism based on the conceptor framework that regulate the RNN dynamics in an adaptive manner towards a predefined reference dynamic. To this end, we establish an adaptive control of the network dynamics, extending its functionality during inference and facilitating the enhanced abilities of RNNs. In particular, RNNs augmented with the conceptor control loop show improved temporal pattern interpolation ability, stabilization against partial network degradation, and robustness against input distortion. 
\end{quotation}

\section{introduction}

Adaptive networks feature a flexible structure that evolves over time and might depend on the state of its internal nodes\cite{berner2023adaptive}. Such adaptive networks find various applications in studying complex dynamical phenomena\cite{gross2006epidemic}, modeling, e.g., power grids~\cite{berner2021adaptive}, learning of neural networks ~\cite{clopath2010connectivity,morales2021unveiling} and in the control of complex systems~\cite{xuan2011structural}. Their adaptive nature allows such networks to dynamically adjust to evolving external conditions and internal dynamics. Such networks are particularly suited for applications requiring robustness against changing environments , with the added capacity to self-organize. Thereby, adaptivity refers to the ability to modify structure and function in response to new information, disturbances, or changing objectives. Such adaptability is achieved through mechanisms that enable the network to reconfigure its connections, adjust its parameters, or even alter its overall architecture in real-time or across different operational phases. 

In the context of machine learning, artificial neural networks are applied to solve regression and classification tasks. In particular, recurrent neural networks (RNNs) have proven remarkably successful in predicting complex dynamics due to their inherent ability to capture temporal dependencies. Accordingly, RNNs have been applied to various tasks such as time series prediction~\cite{hewamalage2021recurrent}, language modeling~\cite{sutskever2011generating}, classification of dynamics~\cite{ganaie2020identification} and control of complex systems~\cite{pan2011model}. Beyond machine learning, RNNs can be seen as a general abstract model of network dynamics, which gave numerous dynamical insights in the natural science study of biological evolution, social networks, and brains~\cite{hensSpatiotemporalSignalPropagation2019}. 
In this context, digital and discrete-time RNNs~\cite{jaeger2004harnessing, lukovsevivcius2009reservoir} are used as well as analog continuous time networks implemented in various hardware can be employed\cite{Appeltant2011, brunner2013parallel, nakajima2021reservoir}. Relying on learning and optimization mechanisms such as backpropagation through time~\cite{Werbos1990} and reservoir computing (RC)~\cite{nakajima2021reservoir, jaeger2001}, certain parts of the network are adapted during the training phase. This adaptation increases the network's performance in the task at hand. Backpropagation through time employs gradient descent methods to update every parameter of the network. On the other hand, RC simplifies the training by only adapting a set of linear output weights. During RNN training, both methods make the network's structure evolve over time to increase performance in a certain task, and hence the network can be interpreted as adaptive. However, after the training, the parameters are usually kept fixed during inference to exploit the learned behavior. This prevents the RNN from continuing to adapt. Recently, various works introduced an extension to reservoir computing framework that allows to control the systems during the inference phase~\cite{klos2020dynamical, Kong2021, kim2021teaching}. Independently, these works proved RNNs effective in learning to forecast complex unseen dynamics given several differently parameterized examples from a dynamical system. By parameterizing certain parameters of the network, they demonstrate enhanced capabilities of RNNs allowing to predict bifurcations, tipping points and transient chaos. Furthermore, exploiting symmetries of delay-dynamical and spatiotemporal systems, adapted network structures facilitate to infer entire bifurcation diagrams unseen during training~\cite{goldmann2022learn,Ji2024learn}. However, similar to standard RC, the parameters of the RNNs are kept fixed rendering them often unable to cope with changing conditions such as changing external signals, changing objectives or degrading aspects of the network 

In this manuscript, we aim to overcome the unadaptive character of RNNs during the inference time by keeping parts of the inner network weights adaptive. We employ the conceptor framework and design an adaptive control loop that allows the RNNs to deal with unforeseen situations. As it will be later shown, conceptors characterize the input-driven dynamics of the networks. A conceptor can be geometrically interpreted as an ellipsoid encapsulating a trajectory given by the network's state evolving over time. Furthermore, using the conceptor, the network dynamics can be controlled by projecting it into a certain subspace thus, restricting the potential states ~\cite{jaeger2014controlling}. To make RNNs adaptivtive, we design a conceptor-based control loop that estimates the network's currently occupied subspace. We then use this subspace as a target to regulate the behavior of the RNN in the presence of unforeseen perturbations. By doing so, we control the RNN dynamics in a flexible way and can increase the network's computational capacity resulting in enhanced interpolation, robustness against partial network failure, and stabilization of input-driven dynamics.

The manuscript is structured as follows: In section~\ref{sec:conceptor}, we show how to control RNNs using the conceptor framework. Subsequently, in section~\ref{sec:ccl}, we employ the conceptor framework to design a control loop that renders the RNN adaptive even in the inference phase. In section~\ref{sec:functionality}, we demonstrate the extended functionality of the adaptive RNN for different scenarios, including interpolation and robustness. Finally, we demonstrate an advanced control setting for input distortion in section~\ref{sec:extending}. Section~\ref{sec:conclusion} concludes the results presented in this manuscript. 

\section{Controlling network dynamics using conceptors\label{sec:conceptor}}
\begin{figure}[!t]
    \centering
    \includegraphics[width=0.49\textwidth]{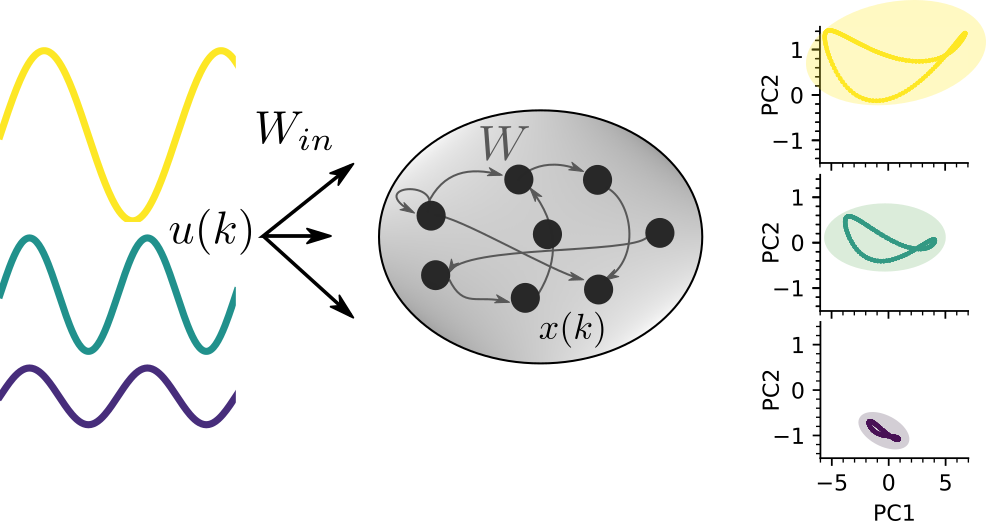}
    \caption{Recurrent neural network subject to three different input time series with different frequencies and amplitudes (color-coded). On the right side, we show the affine projection of the network dynamics related to the three different input patterns onto a low-dimensional manifold using principal component analysis (here using the two largest principal components).\label{fig:neural_geometry}}
\end{figure}

In this section,present a control scheme for the high-dimensional dynamics of RNNs either subject to an external input drive or evolving autonomously. The dynamical behavior of the system here employed, called leaky RNN~\cite{jaeger2001}, is governed by the following discrete-time equation:
\begin{equation}
    x(k+1) = (1-\alpha)x(k) + \alpha \tanh(W x(k) + W_{in} u(k) + b), \label{eq:leaky_rnn}
\end{equation}
where $x(k)\in\mathbb{R}^N$ is the state of the network, $\alpha<1$ is the leak-rate, $W\in\mathbb{R}^{N \times N}$ are the recurrent connections, $W_{in}\in\mathbb{R}$ are the input weights, $u(k)$ is the input signal and $b\in\mathbb{R}^N$ is a bias vector. The $\tanh$ nonlinearity is applied element-wise.

In the following, we drive the leaky RNN with different input sequences $u(k)$ and depending on the parameters of the network and the input signal $u(k)$, the nodes of the network start to oscillate. These nonlinear oscillations in the state space $x(k)$ can be seen as reverberations (analogous to how an object dropped into water causes ripples to spread out and reverberate long after the initial impact\cite{fernandoPatternRecognitionBucket2003}). In Fig.~\ref{fig:neural_geometry}, we plot the reverberations of a RNN independently driven by three different sine-waves represented in a low-dimensional projection using Principal Component Analysis (PCA). Thereby, PCA on the network's state space x(k), finds the axes that capture the most of the variance in decreasing order. In this context, the principal components can be viewed as the axis of an ellipsoid that describes the geometry of the reverberation in the network's state space. Different characteristics of the input signal are encoded into certain trajectories within the network's state space that can be distinguished via their encapsulating ellipsoids (see the shadowed regions in the right part of Fig. 1). Analyzing the eigenvalue spectrum of the principal components thereby exhibits that most of the variance of the RNN’s dynamics appears in a low-dimensional subspace of the network's high-dimensional state space.

The so-called conceptor framework introduced by \citet{jaeger2014controlling} exploits the fact that the dynamics of driven and autonomous are often lying on a low-dimensional subspace. Additionally, when the dynamics are not low dimensional (the correlation matrix is full rank), filtering out the low-variance directions does not damage the rest of the system. The conceptor $C$ of an RNN is defined via the following objective function considering the state $x$ as a random variable: 
\begin{align}
    \mathbb{E}_x\|Cx - x\| + \gamma^{-2}\|C\|_{\text{fro}}^2 \label{eq:obj}
\end{align}
where $\gamma$ is a control parameter called \textit{aperture}, and $\|\cdot\|_{fro}$ indicates the Frobenius norm. Mathematically, the conceptor is a soft-projection matrix $C$ that minimizes the L2-distance between the projected state $Cx$ and the original state $x$ averaged over time. Using a regularization parameter $\gamma$, the second term controls how many dimensions within the network's state space are considered for the projection. The objective function given in Eq.~\ref{eq:obj} has a unique analytical solution given by:
\begin{align}
    C &=  R (R + \gamma^{-2} I)^{-1} \label{eq:conceptor}
\end{align}
where $R= \mathbb{E}_x[x^T x]$ is the $N \times N$ correlation matrix of $x$ and $I$ is the  $N \times N$ identity matrix~\cite{jaeger2014controlling}.

Once computed for a given input sequence, the conceptor $C$ can be used to control the dynamics of the RNN by inserting it into Eq.~\ref{eq:leaky_rnn} yielding:
\begin{equation}
    x(k+1) = C [(1-\alpha) x(k) + \alpha \tanh(Wx(k) + W_{in}u(k) + b) ],
    \label{eq:proj_dyna}
\end{equation}
where the symmetric positive definite conceptor matrix $C$, softly project the dynamics to preserve the relevant dynamical features of $x$. Accordingly, a conceptor can be seen as a geometrical characterization of a trajectory that arises within an evolving RNN; it captures where the trajectory lies. Furthermore, Eq. \ref{eq:proj_dyna} shows that this characterization can also be used for controlling the RNN. Whereas the computation of the conceptor can be carried out after computing the network dynamics driven by an input time series, there exists also an online computable version referred to as \textit{autoconceptor}~\cite{jaeger_using_2017}. Within the autoconceptor framework, the conceptor is recursively updated at every time step $k$ by the following equation: 
\begin{equation}
    C(k+1) = C(k) + \eta(x(k) - C(k)x(k)x^T(k) - \gamma^{-2}C(k)). \label{eq:autoconceptor},
\end{equation}
where $\gamma$ is the aperture of the conceptor, and $\eta$ is the learning rate. While Eq. \ref{{eq:proj_dyna}} had a static (non-varying in time) conceptor biasing the unfolding of the RNN dynamics, here we have a dynamic conceptor that evolve in time and converge to the static conceptor. Note that, the learning rate needs to be set sufficiently small to average over a sufficient number of time steps to converge to the static conceptor. The convergence is guaranteed by the fact that the autoconceptor dynamic (Eq. \label{eq:autoconceptor}) is derived from a stochastic gradient descent procedure on the conceptor loss~\cite{jaeger2014controlling} (Eq. \ref{eq:conceptor}). 

In general, the conceptor framework can be used to control RNNs and perform a variety of information processing tasks such as denoising, multi-tasking and associative memories~\cite{jaeger2014controlling}. Furthermore, one can define multiple operations based on the conceptor matrices, such as a set of logical morphing, interpolation operations, and learning rules to learn incrementally a sequence of tasks. The latter was recently used to improve the continual learning abilities of deep feed-forward networks~\cite{he_overcoming_2018}. 

The conceptor framework enhances the functionality of RNNs, given the network dynamics defined by equation~\ref{eq:proj_dyna}. However, the conceptor does not yet render the network adaptive since it is defined as a static projection. Such nonadaptive control renders systems often unstable and not robust against external perturbations. Therefore, in the next section we propose an extension of the conceptor framework, yielding an adaptive control mechanism for RNNs. Subsequently, we compare computational capabilities of the nonadaptive conceptor only and the proposed adaptive conceptor control loop framework in three challenging tasks, namely temporal pattern interpolation, network degradation and input perturbation. 

\subsection{Adaptive control of network dynamics using a conceptor control loop\label{sec:ccl}}

\begin{figure}[!t]
    \centering
    \includegraphics[width=0.49\textwidth]{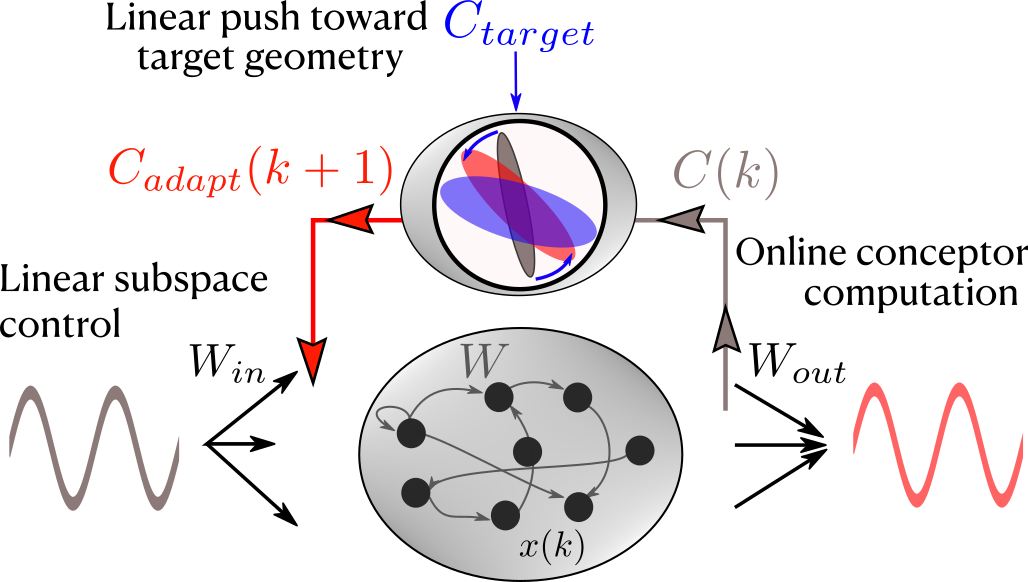}
    \caption[Schematic conceptor control loop]{Conceptor control loop where the current conceptor of the network $C$(k) is measured via the autoconceptor framework and slightly linearly adapted towards the target conceptor $C_{target}$. The adapted conceptor $C_{adapt}(n)$ is then applied to the network. Thereby, the dynamics of the network are pushed towards a linear subspace that are close to the targeted subspace.}
    \label{fig:ccl}
\end{figure}

In this section, we propose to use the system composed of the conceptor and RNN given by Eq. ~\ref{eq:proj_dyna} together with a specifically designed control loop. This conceptor control loop (CCL) aims to balance the control of the dynamics towards predefined target dynamics and the dynamical stability of the network. The CCL is schematically depicted in Fig.~\ref{fig:ccl} encompassing three main steps: 
\begin{enumerate}
    \item estimating the current conceptor $C(k)$ of the RNN in an online fashion from observering the sequence of states $x$
    \item pushing the estimated conceptor into the direction of the predefined target conceptor $C_{target}$,
    \item using the linearly pushed conceptor $C_{adapt}$ to control the network.
\end{enumerate}
In our approach, we do not directly enforce the target conceptor as in Eq.~\ref{eq:conceptor} but rather push the plugged-in conceptor step-by-step toward the target subspace in an adaptive way.

Based on the autoconceptor framework given by Eq.~\ref{eq:autoconceptor}, we derive a control loop that adapts the estimated conceptor toward a target conceptor $C_{target}$ according to:
\begin{align}
    C(k+1) &= C(k) + \eta(x(k) - C(k)x(k)x^T(k) - \gamma^{-2}C(k)) \\
    C_{adapt}(k) &= C(k) - \beta (C(k) -C_{target} ) 
    \label{eq:adaptive}
\end{align}
where $x(n)$ is the network state, $C_{adapt}(k)$ is the adaptive conceptor, $\gamma$ is the aperture of the conceptor and $\beta$ is the gain that determines the strength of the linear push. The target conceptor can be given as a previously defined conceptor or as the mixture of several other conceptors (see ~\ref{sec:interpolation}) depending on the task that needs to be addressed. Finally, the adaptive conceptor is applied to the network using the following equation:
\begin{align}
    x(k+1) =& C_{adapt}(k) [(1-\alpha) x(k) + \alpha \tanh(Wx(k) \nonumber \\ &+ W_{in}u(k) + b) ].
    \label{eq:c_adapt_on_the_loop}
\end{align}

\section{Adaptive control of autonomous RNNs\label{sec:functionality}}

\begin{figure*}[!htb]
    \centering
    \includegraphics[width=0.975\textwidth]{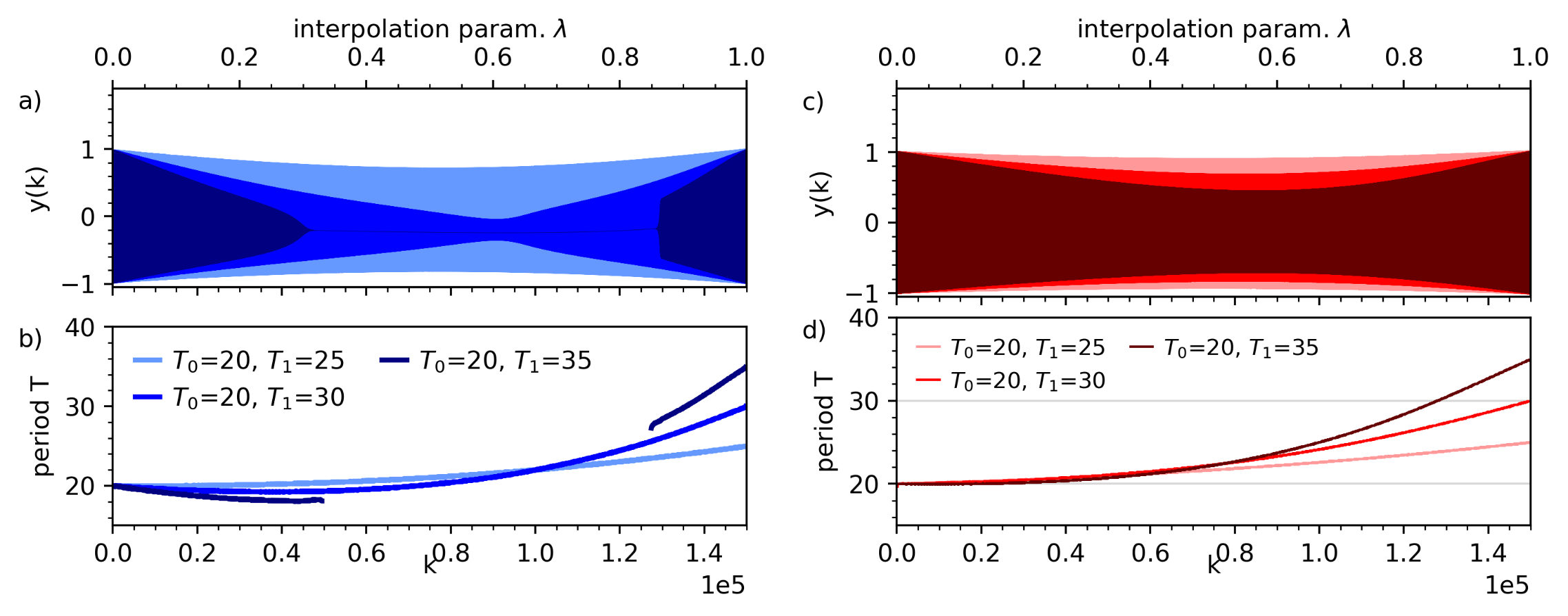}
    \caption{\label{fig:interpolation} a) Autonomously generated output $y(k)$ of the conceptor-RNN trained on two sinusoidal time series with $T_0=20$ and different $T_1\in[25,30,35]$ (different blue shades). Along the x-axis the plugged in conceptor is a linear interpolation between the conceptors $C_0$ and $C_1$ defined during training. b) Period of the intermediary solutions while scanning the interpolation parameter $\lambda$ in the range 0 to 1.  c) Output of the conceptor-RNN with applied conceptor control loop during interpolation between the two conceptors $C_0$ and $C_1$ generated for different periods of the training time series given by two sinusoidal time series with $T_0=20$ and different $T_1\in[25,30,35]$ (different red shades). Similar to a) and b), along the x-axis the interpolation parameter $\lambda$ is scanned in the range 0 to 1. d) Varying period of the autonomously generated time series $y(k)$ during the interpolation while applying the adaptive conceptor control loop.}
\end{figure*}

In this section, we compare the computational capabilities of RNNs either with the nonadaptive conceptor-only framework or with the applied adaptive CCL on two different tasks. For both tasks, the network is trained to reproduce a given time series/temporal pattern without external input. First, we perform interpolation of temporal patterns and. In a second application example, we demonstrate the stabilization of the dynamics against partial network degradation. Here we train the RNN using the RC paradigm to reproduce a temporal pattern $u(k)$ by optimizing a linear set of output weights. Accordingly, the weights are optimized to predict the input $u(k)$ of the RNN one step ahead by setting the prediction target to be $u(k+1)$. Since the output weights are linear, we can apply linear or ridge regression to find the optimal parameters $W_{out}$\cite{jaeger2001, jaeger2004harnessing,nakajima2021reservoir} giving the output $y=W_{out}x$. Furthermore, we determine the spectral radius $\rho$ of the internal weights $W$ and a gain scaling $\rho_{in}$ of the input weights $W_{in}$ using Bayesian optimization~\cite{yperman2016bayesian}. After the training, the RNN is set into the so-called autonomous mode, running solely based on its own predictions. The dynamics of the RNN in the so-called \textit{autonomous mode} are governed by the following equation\cite{jaeger2014controlling}:
\begin{align}
    x(k+1) =& C [(1-\alpha) x(k) \nonumber \\ 
    &+ \alpha \tanh(Wx(k) + W_{in}W_{out}x(k) + b) ], \label{eq:autonomous}
\end{align}
where the conceptor $C$ can be either static and predefined or adaptive as proposed in the CCL given in Eq.~\ref{eq:adaptive}. In this mode there are no external input signals that could guide the network's dynamics. Assuming the training is accurate, feeding back the next-step prediction as input allows the system to regenerate the same trajectory as if it was input-driven. 

\subsection{Interpolation between temporal patterns\label{sec:interpolation}}

In the following, we evaluate the static conceptor and the adaptive CCL in an interpolation task as proposed e.g. by \citet{wyffels2014frequency}. The goal of this task is to generate an intermediate temporal pattern that continuously interpolates features such as the frequency between two time series observed during a training phase. Here, we train a RNN on two sine waves with different frequencies as input data during the training phase. The sine waves are given from the model family as follows:
\begin{equation}
    u_q(n) = \sin(2\pi T_q n) \text{ with } q \in \{0, 1\} , 
\end{equation} .
where $T_0=20$ and $T_1>T_0$. The two sine waves are used to train the network output weights $W_{out}$ for one-step-ahead prediction using ridge regression. Additionally, for each of the two time series a conceptor is computed yielding $C_0$ and $C_1$. After training, we run the network in the autonomous mode given by Eq. ~\ref{eq:autonomous}. 

We start by investigating the ability of the nonadaptive conceptor-RNN to generate intermediate dynamics that interpolate between two temporal patterns. In his work, \citet{jaeger2014controlling} already outlines the ability to transition from one conceptor into another via linear interpolation of the conceptor matrices computed during the training:
\begin{align}
    C_{target}(\lambda) = \lambda C_0 + (1-\lambda) C_1, \label{eq:lin_interp}
\end{align}
where $\lambda$ is the interpolation parameter. By scanning the interpolation
parameter $\lambda$ while running the network in the autonomous mode, the network’s dynamics are projected towards the intermediate conceptor space. This projection allows the network to generate a set of new dynamics in a controlled manner.
In Fig.~\ref{fig:interpolation}, we show the results of three different interpolations between sine waves with increasing deviations from their initial periods. In Fig.~\ref{fig:interpolation} a), we present the output of the RNN while scanning the interpolation parameter $\lambda$ in the range 0 to 1 slowly, i.e. $\lambda=10^{-5}k$. The output states indicate that the network can generate, for most of the conditions, an intermediary sine wave pattern yielding intermediate periods. However, with increasing distance of the trained periods T0 and T1, the amplitude of the intermediary sine wave decreases. For $T_0=20$ and $T_1=35$, the network converges into a fixed point in the range $\lambda\in[0.35,0.85]$ so no oscillations can be identified. In this range, the control of the RNN fails, and the interpolation breaks down. In Fig.~\ref{fig:interpolation} b), we analyze the period of the intermediary solutions, where we find that the period of the intermediary solutions varies along the interpolation parameter as expected and hence yields increasing periods while transitioning from $\lambda=0$ towards $\lambda=1$. Accordingly, the static conceptor only framework might allow for the interpolation of close-by temporal patterns. However, it is not sufficient to generate intermediary dynamics especially if the periods of the two input patterns are far apart due to fixed point dynamics. 

The reasons for the breakdown of interpolation when using the static conceptor framework might be twofold. Firstly, the interpolated conceptor enforces dynamics in a linear subspace of the network's state space that was not observed during training. Hence the dynamics within that subspace can not directly be optimized to be stable. Furthermore, the output weights are trained only on the two initial time series, whereas we apply them to intermediate dynamics that can be far apart from the training dynamics. Secondly, the linear interpolation of the conceptor matrix introduced in Eq.~\ref{eq:lin_interp} is known to shrink the eigenvalues of the resulting interpolated conceptor\cite{schultz_geometries_2017}. This shrinkage of the eigenvalues, known for symmetric positive definite matrices, might lead to a loss of information and in turn, might be one reason for the decreasing amplitudes of the intermediary solutions. As discussed by \citet{schultz_geometries_2017}, the shrinkage of the eigenvalues can be avoided by using another geodesic metric to derive the interpolation scheme, e.g. log-euclidean, affine invariant, or the shape and orientation rotation metric. However, these metrics, due to their reliance on matrix exponentials, are computationally expensive and often numerically unstable~\cite{schultz_geometries_2017}. 

We now proceed to test the ability of the adaptive CCL for interpolation between the two temporal patterns as carried out above. Therefore, we use the same network parameters (see Tab. \ref{tab:hyperara}) as used for Fig.~\ref{fig:interpolation} a) and b) and train the network on the same set of sine waves as above. After training, we apply the CCL to the network and scan the interpolation parameter $\lambda$ in the range 0 to 1. In Fig.~\ref{fig:interpolation} c) and d), we show the output of the network along this scan. We obtain that the network can generate an intermediary sine wave pattern for the three sine wave patterns $T_1\in[25,30,35]$. In contrast to the conceptor-only framework, the adaptive CCL achieves a much more stable oscillation amplitude throughout all three interpolations. Additionally, the periods of the interpolated temporal pattern exhibit a smooth and constantly increasing transition from the starting frequency $T_0$ towards $T_1$ as shown in Fig.~\ref{fig:interpolation} d). Accordingly, the adaptive character of the CCL results in a interpolation much more stable and consistent interpolation than the static scheme. 

\begin{figure}[!htb]
    \centering
    \includegraphics[width=0.49\textwidth]{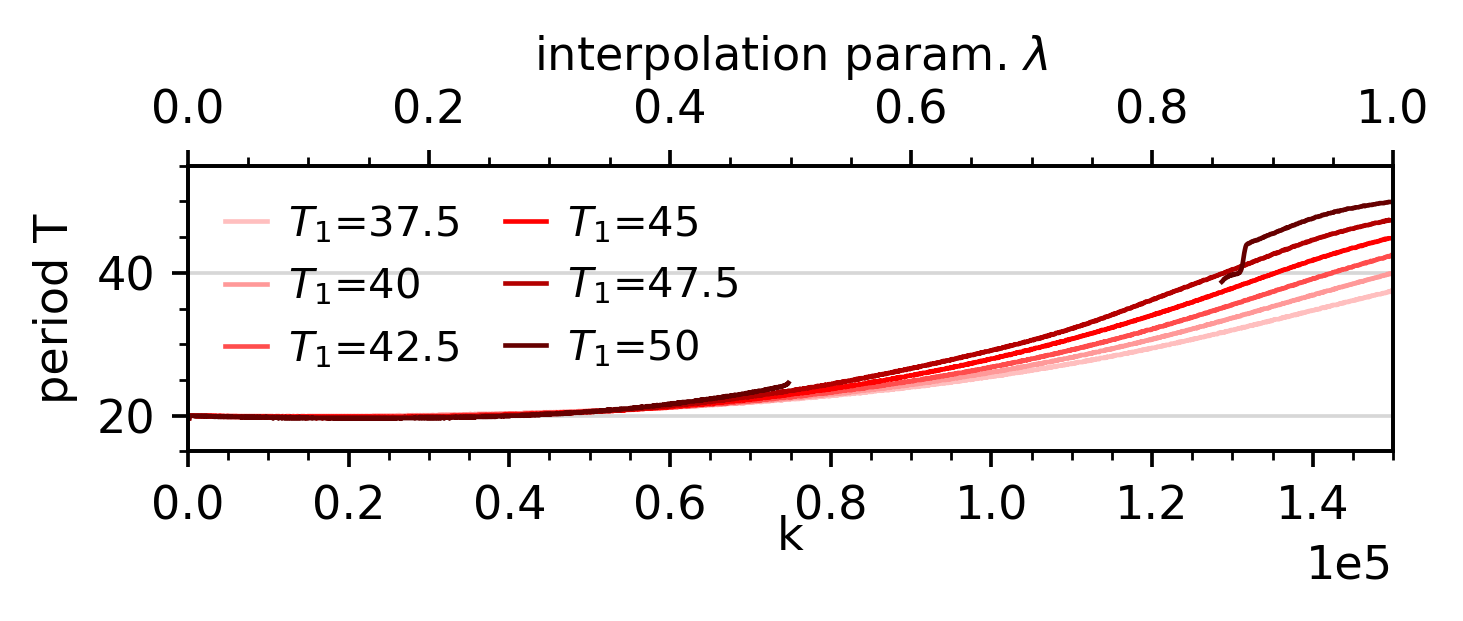}
    \caption{Period of the output patterns  generated by a RNN with applied conceptor control loop to interpolate sine wave time series with different input periods $T_0=20$ and $T_1\in[37.5,50]$.}
    \label{fig:extend_period}
\end{figure}

Moreover, additional experiments shows that the CCL can further boost the ability of the RNN to even interpolate between temporal patterns of more distant periods (Fig.~\ref{fig:extend_period}). Whereas, the conceptor-only framework breaks down to interpolate time series with periods $T_0=20$ and $T_1=35$, here, we demonstrate continuous interpolation of the CCL of periods from $T_0=20$ up to more than twice the initial period at $T_1=47.5$ avoiding fixed point solutions in between. Accordingly, we argue that the proposed adaptive CCL improves the interpolation abilities of the RNN  beyond the static-conceptor-only framework by enhancing the generalization ability of the network without further training. The interpolation capabilities of the CCL eventually break down for a longer period of $T1=50$. In this case, an additional input example in the training set with a period in between $T0$ and $T1$ would extend the interpolation capabilities. We note that the CCL makes a more optimal use of the input examples than the conceptor-only framework.   

From a machine learning perspective where the objective is to generalize from a few samples, the CCL can be seen as a way to adaptively enforce a prior at inference time on how to generalize. The prior consists in assuming that intermediary samples are generated by dynamics with intermediate ellipsoids specified by the linear interpolation between conceptor. With the CCL, we find that the interpolated dynamics generating the prediction lies in intermediary ellipsoids and facilitate a strong similarity with the ellipsoids elicited by the training data. In the case of the static conceptor where the interpolation leads to fixed point dynamic, the ellipsoid of the dynamic is reduce to a point. In this case the prior is violated.

\begin{table*}[!htb]
    \centering
    \label{tab:my_label}
    \begin{tabularx}{0.975\textwidth}{X X |X |X |X}
        \toprule
        {Parameter Name} & {Symbol} & {Interpolation} & {Network Degradation} & {Input Distortion} \\
        \hline
        Spectral radius & \(\rho\) & 1.6 & 0.749 & 0.9\\
        Input scaling & \(\rho_{in}\) & 1. & 1.149 & 0.9\\
        Bias scaling & \(\rho_{b}\) & 1. & 1.5 & 0.2\\
        Leakage & \(\alpha\) & 0.75 & 0.988 & 1.0\\
        Ridge regularization & \(\lambda\) & 0.0001 & 1000 & 0.01\\
        Aperture & \(\gamma\) & 25  & 31.6 & 8\\
        Network size & \(N\) & 256 & 1500 & 50\\
        Learning rate & \(\eta\) & 0.2 & 0.001 & 0.8\\
        Control gain & \(\beta\) & 2.5e-5 & 0.7 & 4e-3\\
        Random Feature Conceptor size & \(N_{RFC}\) & / & / & 200\\ 
        \bottomrule
    \end{tabularx}
    \caption{Hyperparameters used for the different experiments.}
    \label{tab:hyperara}
\end{table*}

\subsection{Stabilization against partial network degradation \label{sec:degradation}}

\begin{figure*}[!htb]
    \centering
    \includegraphics[width=0.9\textwidth]{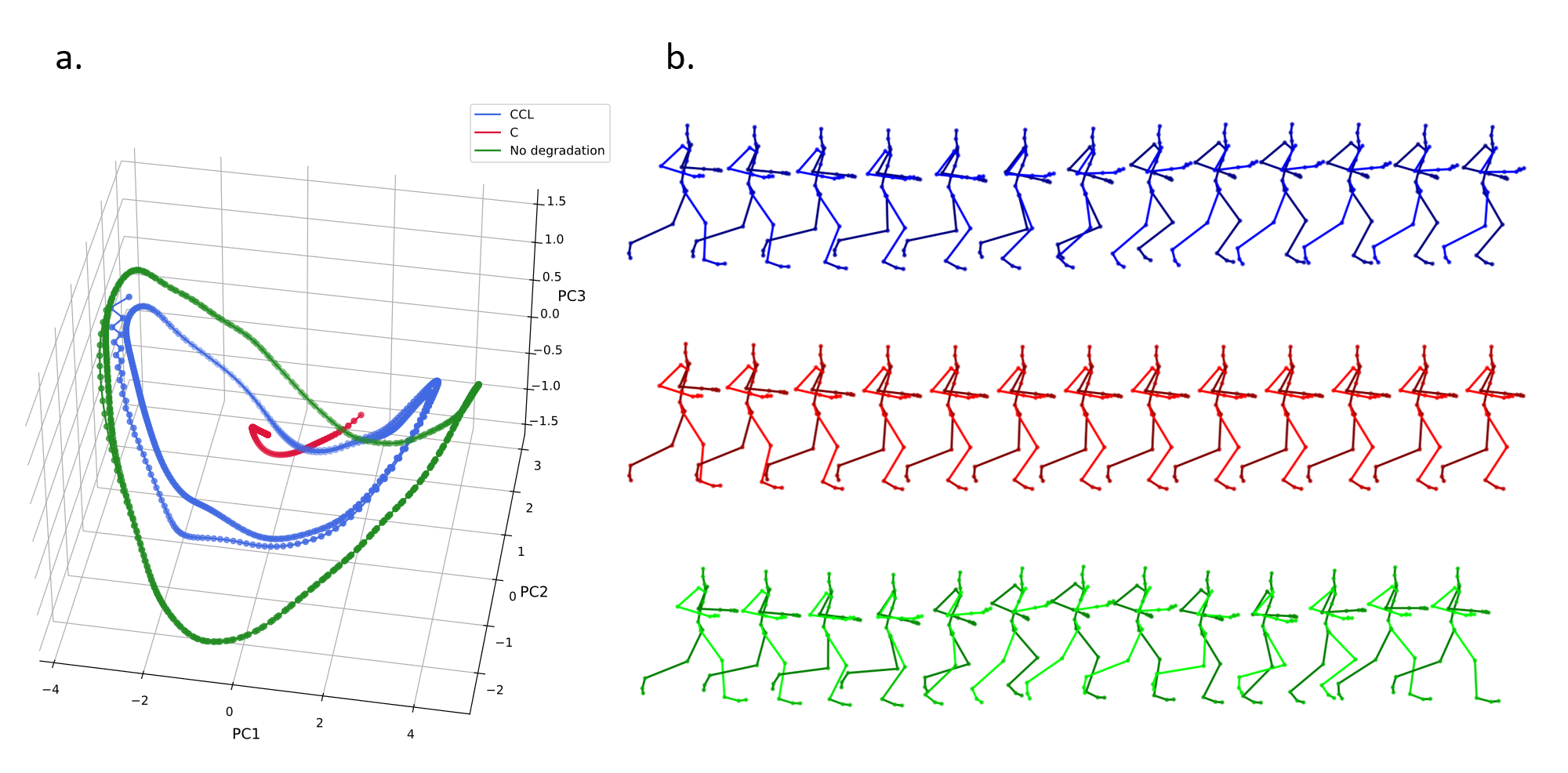}
    \caption{a) Projection of the output of the RNN trained to predict a 94-dimensional time series motion capture time series of a human running behavior. The RNN is degraded by removing a random subset of 200 of its 1500 neurons. The trajectory is depicted for the RNN with the conceptor control loop (blue), compared with only the static conceptor (red), and with a baseline where the RNN is not degraded (green) along the three first principal components of the output. b) The same output generation (same color code) is represented with the geometry of the human body angles. The number of steps depicted is the same for all the conditions and is chosen to showcase one full cycle of the baseline (green).}
    \label{fig:net_degradation}
\end{figure*}

\begin{figure}[!htb]
    \center
    \includegraphics[width=0.49\textwidth]{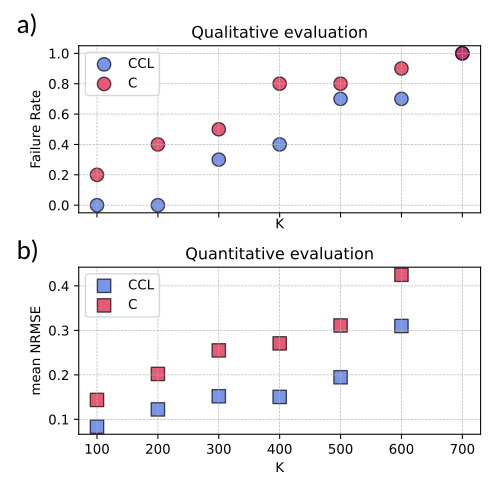}
    \caption{a) Qualitative evaluation of the RNN's ability to preserve its memorized periodic cycle after the removal of up to $K$ of its 1500 neurons. For each increment in $K$, ten distinct trials were conducted with randomly selected neurons removed. The qualitative failure rate is determined via a threshold of 1.0 on the variance of the principal component of the RNN. b) Quantitative assessment of the performance by computing the mean Normalized Root Mean Squared Error (NRMSE) in comparison with the baseline (non-degraded RNN output) after phase alignment over the trials that are non-failing for both conditions (CCL and C). The NRMSE was not computed for $K=700$ as all trials failed to sustain a periodic cycle}
    \label{fig:quant_qual}
\end{figure}

In this section, we evaluate the capabilities of the static conceptor only and the adaptive CCL to enhance the robustness of networks to partial degradation. Partial degradation refers to the process of removing neurons from the network in the inference while evaluating how well the network is able to continue performing the task it learned during the training. Historically, psychologists and neuroscientists have recognized that artificial and biological neural networks share a characteristic known as "graceful degradation"~\cite{mcclellandParallelDistributedProcessing1986}. This term describes how a system's performance declines gradually as more neural units are compromised, without a singular, catastrophic failure point. Inspired by early interest in this phenomenon, numerous engineering strategies have been developed to leverage and augment the inherent fault tolerance in ANNs~\cite{torres-huitzilFaultErrorTolerance2017}. Most of these strategies lack the ability to adapt dynamically. Examples of such strategies include the injection of noise into neuron activity during training, the addition of
regularization terms to improve robustness, or the design of networks with built-in redundancy by duplicating important neurons and/or synapses. However, research on improving graceful degradation within RNNs, especially those handling dynamic inputs, remains sparse~\cite{torres-huitzilFaultErrorTolerance2017}.

To study the influence of degradation within RNNs, here we train the output weights of a network in one step ahead prediction. We then feed the prediction back to generate a pattern autonomously as defined by Eq.~\ref{eq:autonomous}. As the target time series, here we use human motion capture data from the MoCap data set~\cite{CMUMocap}. The corresponding time series provides the data of 94 sensors attached to a human body sampling the movement of joints and limbs in a three-dimensional space. For the training, we use 161-time steps that include two periods of a running behavior as shown in Fig.~\ref{fig:net_degradation}. After optimizing the weights and parameters of the RNN for the autonomous continuation (see Tab. \ref{tab:hyperara}), we start to degrade the network by removing $R$ randomly selected neurons from the network. More specifically, we clamp the corresponding state-coordinate $x$ of the selected neuron to zero while computing the network's autonomous continuation. Finally, we analyze the ability of the network to continue its autonomous prediction despite the partial degradation by comparing the mean square error (MSE) of its output compared to the original time series. In our implementation, we use an RNN of $N=1500$ neurons and first evaluate the robustness of the static conceptor only framework by degrading $R=200$ randomly selected neurons. In  Fig.~\ref{fig:net_degradation} a), we plot the three first Principal Components (PCs) of the autonomously evolving RNN in red. The projection of the evolution of the RNN without degradation is shown for comparison (green color). Similarly to the task of interpolation as described in Sec.~\ref{sec:interpolation}, the applied degradation pushes the system outside the dynamical regime it was trained on. In analogy to the interpolation results, we obtain that the RNN with the applied static conceptor tends to fall into fixed-point dynamics (red curve) rendering the network unable to continue its learned dynamics. In this case, the stickman, whose movements mirror the time series data, becomes unnaturally still, as illustrated in Fig.~\ref{fig:net_degradation} b). With the degradation, the network's output does not produce a running motion, leaving the stickman frozen. Statistically, we find around half of the ten the 10 random degradations we tested to lead to such catastrophic failure. 

We now proceed to evaluate the graceful degradation abilities of the RNN while applying the CCL that we slightly adapted for this challenging task (see App. \ref{degradation_app}).  As shown in Fig.~\ref{fig:net_degradation}, adding the CCL stabilizes the system around a limit cycle after a short transient (blue curve). Across different experiments with a growing number $K$ of degraded neurons, we observe that the CCL consistently stabilizes the system against partial degradation. This adaptation enables the system to maintain its dynamics and qualitatively preserve the limit cycle (Fig. \ref{fig:quant_qual}a)). From visual inspection, we can also see that the CCL is able to preserve appealing properties of running as shown in Fig.~\ref{fig:net_degradation}b). Note that the recovery from the degradation is only approximate, and the stickman from the CCL system is slightly slowed down compared to the baseline (green), which achieves one full cycle in a shorter time. 

In our quantitative evaluation, we assessed the continuation capabilities of the non-failing trials by comparing the time series from the baseline (non-degraded RNN) with the phase-aligned output from the degraded RNN using Normalized Root Mean Squared Error (NRMSE). The results show that the average NRMSE is consistently lower for the system incorporating the CCL (Fig.~\ref{fig:quant_qual}b)). Overall, both quantitative and qualitative results highlight that the adaptivity endowed by the CCL improves the robustness of the RNN against network degradation. Here, we show that the CCL is a straightforward, yet effective enhancement allowing for graceful degradation of internal neurons.

% without requiring more information about the system (jacobian...) 

\section{Extending Conceptor to non-autonomous systems - A hierarchical CCL architecture to increase robustness against input distortion \label{sec:extending}}%Homeo-dynamics
\subsection{Robustness against input distortions}

In the previous section~\ref{sec:functionality}, we evaluated two different tasks setting up the RNN to continue autonomously a certain temporal pattern. In contrast, in this section, we will evaluate the capacity of an input-driven RNN to continue to do its task even after its input gets perturbed. The task we aim to solve within this section contains the adaptation of the network to equalize an input signal that, compared to the training, is corrupted by an unpredictable distortion, e.g., a strong reduction in the signal amplitude. Note that despite the transformation being linear, the RNN response to this change is nonlinear. In signal processing terms, we are looking for a mechanism of adaptive blind nonlinear channel equalization\cite{farhang-boroujenyAdaptiveFiltersTheory2013}. This task may be related to cognitive science and cybernetics where biological agents are endowed with homeostatic mechanisms to be \textit{ultrastable}\cite{ashbyDesignBrainOrigin2014} against environmental perturbation or when recovering transmitted signals that were intentionally hidden for secure communications~\cite{fischer2000synchronization}. In a related example to understand human behavior, Kohler \cite{kohlerFormationTransformationPerceptual1963} studied subjects using inverting goggles and demonstrated human adaptability to unpredictable (not seen during ontogeny nor phylogeny) significant perceptual shifts, suggesting a specialized mechanism for managing a perturbation of dynamical visuomotor coordination. Previous works  have attempted to simulate this adaptability, with limited success in explicability and robustness \cite{diapaoloHomeo}. Our work is an attempt in the direction to embed further adaptability properties to RNNs and explain the robustness of biological neural networks.

For this robustness task, the dynamics of the RNN are governed by Eq. \ref{eq:c_adapt_on_the_loop} that we repeat here for clarity:
\begin{align}
    x(k+1) =& C_{adapt}(k) [(1-\alpha) x(k) \nonumber \\ 
    &+ \alpha \tanh(Wx(k) + W_{in}u(k) + b) ].
\end{align}
During training, the system learns the prediction target from a single presentation of the undistorted input time series. Similar to the previous tasks described in Sec. \ref{sec:functionality}, we employ linear regression for next-step prediction and compute a single target conceptor $C_{target}$. In the inference phase, the system aims to counteract a distortion applied to the input series not present during the training.  

\subsection{Adapting the conceptor control loop for non-autonomous systems \label{sec:rfc_apdx}}

\begin{figure}[!htb]
    \centering
    \includegraphics[width=0.49\textwidth]{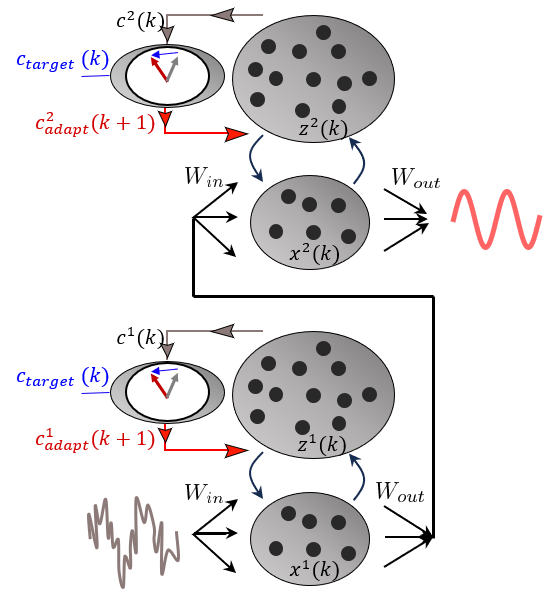}
    \caption{Hierarchical architecture of two layers of RNN. Each layer is composed of a RNN with two sub-layers where the $z^i$ part corresponds to a random expansion into a higher dimension that allows the simplification of the conceptor computation to vector computations (the ellipsoids are replaced by vectors).
    The RNN parameters and the conceptor reference are the same in the two layers. The distorted input is injected at at the bottom of the hierarchy and the undistorted input is at the output of the last layer. }
    \label{fig:architecture}
\end{figure}

\begin{figure*}[!htb]
    \centering
    \includegraphics[width=0.975\textwidth]{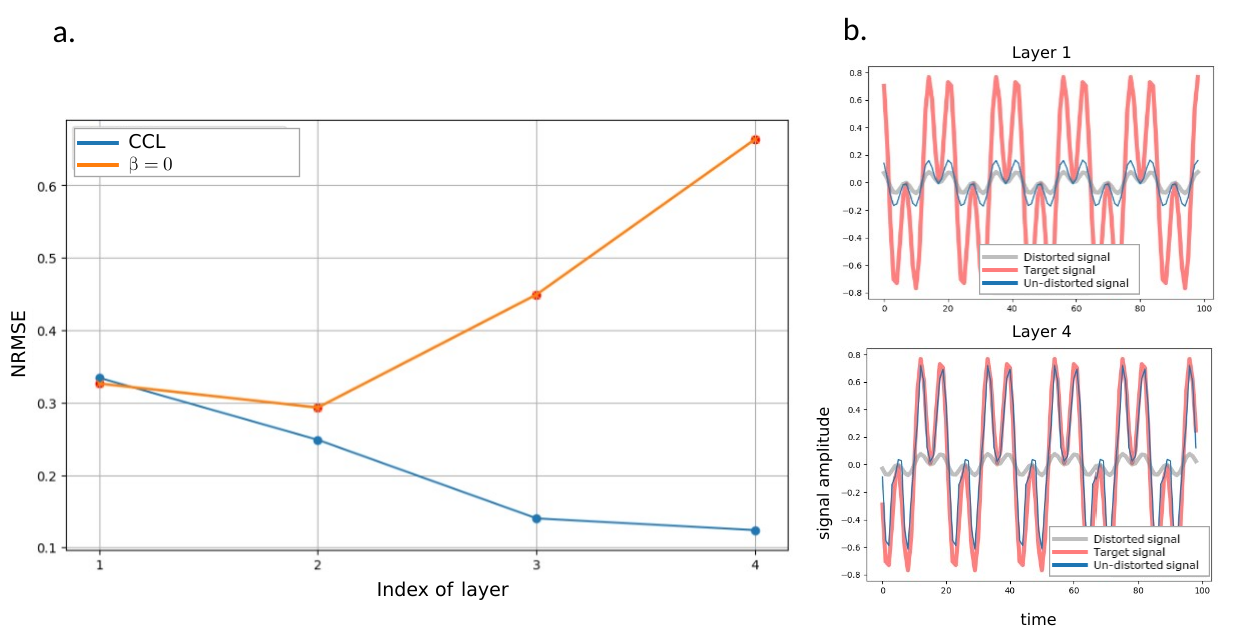}
    \caption{Robustness to strong scaling distortion. a) NRMSE of the output for the different layers in the hierarchy of the architecture. b) Time series output of the first and last layer of the hierarchy.}
    \label{fig:input_distortion}
\end{figure*}
Adapting the way an input-driven system processes information is difficult. This is because changes to the input can push the system away from how it is supposed to function. Conceptually, this altered input acts akin to a signal from an unseen dynamical system coupled with the RNN. Hence, it closely parallels the challenge of regulating a coupled autonomous system where only a fraction of the entire system is accessible for observation and control.

Here we employ a hierarchical architecture of a multi-layer system as illustrated in Fig. \ref{fig:architecture}. Within the deep architecture, we replicate the trained RNN  $L$-times, organizing these replicas into a hierarchical structure of $L$ layers, with the lowest layer connected to the input. The system's output is derived from the topmost layer. Each layer incorporates its own CCL. Similar to the previous sections, each CCL adapts the network dynamics towards the target conceptor determined during the training phase. Additionally, we utilize a modified RNN-conceptor system referred to as \textit{random feature conceptor} \cite{jaeger2014controlling} wherein the RNN of each layer is split into two sub-layers $x^i$ and $z^i$. The sublayer $z^i$ is generated via a random expansion of the states $x^i$ into a higher dimensional space. This random expansion was shown to allow the replacement of conceptor-related matrix computation by less costly vector-based ones\cite{jaeger2014controlling}. The mathematical description of this architecture is presented in App. \ref{sec:rfc_apdx}.

The CCL of each layer pushes the dynamics toward its reference linear subspace via the unperturbed signal's reference conceptor. This push makes each layer transform its input into an output closer to the unperturbed signal. As shown in Fig.~\ref{fig:input_distortion} a), the hierarchically arranged system progressively eliminates the distortion, reducing the prediction error layer by layer. As a demonstration of the validity of our architecture, we studied a simple four-layer RNN subjected subject to a 0.3 downscaling distortion on a periodic time series composed of two sine waves with periods of 7 and 21. We observe a significant improvement of the accuracy in the time series reconstruction by the fourth layer (Fig.\ref{fig:input_distortion}c), compared to the input-connected initial layers (Fig.\ref{fig:input_distortion}b). We also determine the reconstruction loss characterized by the NRMSE, which shows a consistent decrease across the hierarchy (Fig. \ref{fig:input_distortion}a). We also verify the CCL's causal role in signal undistortion by demonstrating the hierarchical system's inability to correct distortions without the CCL as shown in Fig.\ref{fig:input_distortion} a) (orange curve). This further underlines the CCL's computational capabilities in controlling the hierarchical architectures and handling novel data during inference.

\section{Conclusion\label{sec:conclusion}}

In this manuscript, we introduced an adaptive control loop based on the conceptor framework to augment the learning capabilities of RNNs. We demonstrated that by controlling RNN dynamics during inference, we achieve several significant benefits. First, our method improves the generalization abilities of RNNs, enabling the interpolation of more distinct temporal patterns. Second, it enhances graceful degradation, making RNNs more robust to errors or unexpected conditions. Finally, hierarchical RNN structures with our adaptive control loop exhibit improved signal reconstruction capabilities.

Our findings suggest that maintaining adaptive elements within a network post-training can significantly expand the potential applications of RNNs. This approach has particular promise for enabling RNNs to operate reliably in challenging environments or adapt to novel tasks with minimal additional training (few-shot learning).

While our study focused on recurrent networks, the conceptor framework has the potential to control and augment various dynamical systems, including feedforward and other artificial neural networks. This versatility establishes our work as a foundation for broader exploration into the control and enhancement of nonlinear computing systems.

\appendix
\section*{Supplementary materials}
Supplementary materials for this study include videos that demonstrate the effects of the Conceptor Control Loop (CCL) on RNN behavior. Available videos are 'running.gif' for the CMU 016 55 MoCap data series, 'runningCcl7.gif' showing the seventh example from Fig.\ref{fig:samples_degradation}c) with CCL, and 'runningNoCcl7.gif' without CCL from Fig.\ref{fig:samples_degradation}b). These visuals highlight the RNN's adaptive performance under various conditions, supporting our findings on the benefits of incorporating adaptivity into RNNs.

\section*{Acknowledgement}
The authors acknowledge the support of the Spanish State Research Agency, through the Severo Ochoa and María de Maeztu Program for Centers and Units of Excellence in R\&D (CEX2021-001164-M),  the INFOLANET projects (PID2022-139409NB-I00) funded by the MCIN/AEI/10.13039/501100011033 and through the QUARESC project (PID2019-109094GB-C21 and -C22/ AEI / 10.13039/501100011033) and the DECAPH project (PID2019-111537GB-C21 and -C22/ AEI / 10.13039/501100011033). All authors acknowledge financial support by the European Union’s Horizon 2020 research and innovation program under the Marie Skłodowska-Curie grant agreement No. 860360 (POST DIGITAL). 

\section{Random Feature Conceptor\label{sec:rfc}}
In section \ref{sec:extending} we adapted the RNN and CCL based on the ideas of Random Feature Conceptor (RFC) \cite{jaeger2014controlling} to enhance the system's computational efficiency. This was achieved by substituting adding an expansion in high dimension $z$ where the matrix-based computation of conceptors can be replaced by vector-based analog. Other than for making the system more efficient to implement by replacing vector conceptor to diagonal conceptor. Other than for computational efficiency on a digital computer, this method is more bio-plausible and can be more easily be implemented in neuromorphic hardware\cite{jaeger2014controlling}. The equation of the RNN is now: 
\begin{align}
    x(k+1) &= \tanh(Gz(k) + W^{in}u(k) + b) \\
    z(k+1) &= c_{adapt}(k)*F'x(k+1)
\end{align}
where $F'$ is a random expansion in the $z-$space of higher dimension $N_{RFC}$ and $G$ corresponds to a matrix product between a random compression matrix to come back to the original dimension of $x$ and the weight matrix $W$ of the RNN. Now, the conceptor is a vector of dimension $N_{RFC}$, which is introduced in the RNN dynamic through an elementwise multiplication. Theoretical arguments and experimental evidence show that this vector operation is very close to applying a full matrix conceptor on the non-expanded space \cite{jaeger2014controlling}. Hence RFC can still be interpreted as a computational trick with the same semantic of projecting the dynamic of $x$.
The CCL is now applicable in vector form:
\begin{align}
    c(k+1) &= c(k) + \eta(z(k)^2 - c(k)*z(k)^2) - \gamma^{-2}c(k), \\
    c_{adapt}(k) &= c(k) - \beta(c_{target}-c(k)),
\end{align}
where the autoconceptor dynamics correspond to stochastic gradient descent on a vector-based conceptor objective function analog of Eq.\ref{eq:obj}. 
For each element of $c_i$ of $c$:
\begin{align}
    \mathbb{E}_{z_i}[(z_i-c_iz_i)^2] + \alpha^{-2}c_i^2 
\end{align}

Similarly to Section \ref{sec:degradation}, we've found that mixing the $c$ and $c_{adapt}$ improves performance, so we used the following update for our simulations:
\begin{align}
    c_{adapt}(k+1) =& c_{adapt}(k) + \eta(z(k)^2 - c_{adapt}(k)*z(k)^2),  \nonumber \\
    &- \gamma^{-2}c_{adapt}(k) - \beta(c_{target}-c_{adapt}(k)).
\end{align}
We finally note that the autoconceptor dynamics are favorable. Notably, it's much faster converging because it can be used with a larger learning rate \cite{jaeger2014controlling} (see table \ref{tab:hyperara}).

\section{Network degradation: CCL adaptation and further results \label{degradation_app}}
In our simulation, we've found that slightly adapting the CCL by mixing the computations of the estimated conceptor $C$ and $C_{adapt}$ improved the performance of the system. Hence, the simulation for Section \ref{sec:degradation} was done with the following CCL:

\begin{align}
    C_{adapt}(k+1) = C_{adapt}(k) + \eta(x(k) - C_{adapt}(k)x(k)x^T(k), \nonumber \\ - \gamma^{-2}C_{adapt}(k))  -  \beta (C_{adapt}(k) -C_{target} ) . \label{eq:autoconceptor},
\end{align}
where now there is a single conceptor that both estimates the linear subspace of the current dynamic and integrates the feedback from the target.

\begin{figure*}
    \centering
    \begin{adjustbox}{angle=90}
        \begin{minipage}{\textheight} % Use \textwidth for the size before rotation
            \includegraphics[width=0.975\textwidth]{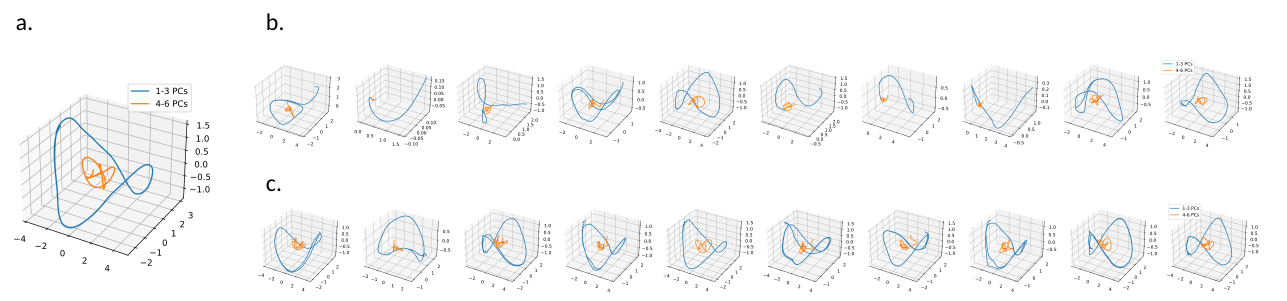}
            \caption{a) The first six principal components of the RNN without degradation. The same Figure but for different samples of random degradation of 200 of the 1500 neurons with (c) and without (b) the CCL.}            
            \label{fig:samples_degradation}
        \end{minipage}
    \end{adjustbox}
\end{figure*}

In Fig.\ref{fig:samples_degradation}b,c), we present further results of the individual trials with different random perturbations of Fig.\ref{fig:net_degradation}b) and for reference the case where the RNN works without degradation Fig.\ref{fig:samples_degradation}a). We also provide videos in the supplementary materials of the running time series CMU 016 55 of the MoCap data (running.gif), and the seventh example of Fig.\ref{fig:samples_degradation} with (runningCcl7.gif) and without (runningNoCcl7.gif) the CCL. The time series were pre-processed as described by \citet{jaeger_using_2017}.

\bibliography{main}% Produces the bibliography via BibTeX.

\end{document}